\documentclass{article}

\PassOptionsToPackage{numbers, compress}{natbib}
\usepackage[final]{template}

\usepackage[pdftex]{graphicx}
\graphicspath{{./figures/}}

\usepackage[utf8]{inputenc} 
\usepackage[T1]{fontenc}    
\usepackage{hyperref}       
\usepackage{url}            
\usepackage{booktabs}       
\usepackage{amsfonts}       
\usepackage{amsmath}       
\usepackage{nicefrac}       
\usepackage{microtype}      

\usepackage{bibunits}
\defaultbibliographystyle{unsrtnat}
\defaultbibliography{extracted}

\usepackage{amssymb}
\usepackage[textsize=tiny, backgroundcolor=yellow, linecolor=black!50]{todonotes}

\def\EE{{\mathbb E}}

\def\RR{{\mathbb R}}
\def\PP{{\mathbb P}}

\def\x{{\mathbf x}}

\def\A{{\mathbf A}}

\def\D{{\mathbf D}}

\def\M{{\mathbf M}}

\def\X{{\mathbf X}}

\def\t{{\mathbf t}}

\def\b{{\mathbf b}}

\def\x{{\mathbf x}}
\def\W{{\mathbf W}}
\def\y{{\mathbf y}}
\def\p{{\mathbf p}}

\newcommand{\bTheta}{\boldsymbol{\Theta}}

\newcommand{\T}{\top}

\usepackage{color}
\definecolor{dred}{rgb}{0.8,0,0}
\definecolor{dgreen}{rgb}{0,0.8,0}
\definecolor{dblue}{rgb}{0,0,0.8}
\definecolor{dpurple}{rgb}{0.8,0,0.8}

\usepackage{paralist}
\renewenvironment{itemize}[1]{\begin{compactitem}#1}{\end{compactitem}}
\renewenvironment{enumerate}[1]{\begin{compactenum}#1}{\end{compactenum}}

\title{Learning Neural Representations of\\ Human Cognition
across Many fMRI Studies}

\author{
Arthur Mensch\footnotemark[1] \\
Inria \\
\texttt{arthur.mensch@m4x.org}
\AND
Julien Mairal\footnotemark[2] \\
Inria \\
\texttt{julien.mairal@inria.fr}
\And
Danilo Bzdok \\
Department of Psychiatry, RWTH \\
\texttt{danilo.bzdok@rwth-aachen.de} \\
\And
Bertrand Thirion\footnotemark[1] \\
Inria\\
\texttt{bertrand.thirion@inria.fr}
\And
Ga\"el Varoquaux\footnotemark[1] \\
Inria\\
\texttt{gael.varoquaux@inria.fr}
}

\begin{document}

\renewcommand*{\thefootnote}{\fnsymbol{footnote}}
\footnotetext[1]{Inria, CEA, Université Paris-Saclay, 91191 Gif sur Yvette, France}
\footnotetext[2]{Univ. Grenoble Alpes, Inria, CNRS, Grenoble INP, LJK, 38000 Grenoble, France}
\renewcommand*{\thefootnote}{\arabic{footnote}}
\setcounter{footnote}{0}

\maketitle

\begin{abstract}
Cognitive neuroscience is enjoying rapid increase
in extensive public brain-imaging datasets. It opens the
door to large-scale statistical models.
Finding a unified perspective for all
available data calls for scalable and automated solutions to an old challenge: how to aggregate heterogeneous
information on brain function into a universal cognitive system that relates mental operations/cognitive processes/psychological tasks to brain networks? We cast this challenge in a machine-learning approach to predict
conditions from statistical brain maps across different studies.
For this,
we leverage multi-task learning and multi-scale dimension
reduction to learn low-dimensional representations of brain images that carry
cognitive information and can be robustly associated with psychological
stimuli.
Our multi-dataset classification model achieves the best prediction performance on
several large reference datasets, compared to models without
cognitive-aware low-dimension representations; it brings a substantial
performance boost to the analysis of small datasets, and can be introspected to
identify universal template cognitive concepts.
\end{abstract}

\begin{bibunit}



Due to the advent of functional brain-imaging technologies,
cognitive neuroscience is
accumulating quantitative maps of neural activity responses to
specific tasks or stimuli.
A rapidly increasing number of neuroimaging studies are publicly shared
(\textit{e.g.,}
the human connectome project, HCP~\cite{vanessen_human_2012}),
opening the door to applying large-scale statistical approaches~\cite{poldrack_scanning_2017}.
Yet, it remains a major challenge to formally
extract structured knowledge from heterogeneous neuroscience repositories.
As stressed in~\cite{newell_you_1973}, aggregating
knowledge across cognitive neuroscience experiments is intrinsically
difficult due to
the diverse nature of the hypotheses and conclusions of the
investigators.
Cognitive neuroscience experiments aim at
isolating brain effects underlying specific psychological processes: they
yield statistical maps of brain activity that measure
the neural responses to
carefully designed stimulus.
Unfortunately, neither regional brain responses nor experimental
stimuli can be considered to be \textit{atomic}: a given
experimental stimulus recruits a spatially distributed
set of brain regions \cite{medaglia_cognitive_2015}, while each
brain region is observed to react to diverse stimuli.
Taking advantage of the resulting data richness to build formal models
describing psychological processes requires to describe each cognitive
conclusion on a common basis for brain response and experimental study
design.
Uncovering \textit{atomic basis functions} that capture the neural
building blocks underlying cognitive processes is therefore
a primary goal of neuroscience~\cite{barrett_future_2009}, for which we
propose a new data-driven approach.

Several statistical approaches have been
proposed to tackle the problem of knowledge aggregation in
functional imaging. A first set of approaches relies on
coordinate-based meta-analysis to define
robust neural correlates of cognitive processes:
those are extracted from the descriptions of experiments --- based on
categories defined by
text mining~\cite{yarkoni_large-scale_2011} or
expert~\cite{laird_brainmap_2005}--- and correlated
with brain coordinates related to these experiments.
Although quantitative meta-analysis techniques provide
useful summaries of the existing
literature, they are hindered
by label noise in the experiment descriptions,
and weak information on brain activation as the maps are reduced to a few
coordinates~\cite{salimi-khorshidi_meta-analysis_2009}.
%
A second, more recent set of approaches models directly brain maps across
studies, either focusing on studies on similar cognitive processes
\cite{wager_fmri-based_2013}, or tackling the entire scope of
cognition~\cite{schwartz_mapping_2013, koyejo_decoding_2013}.
Decoding, \emph{i.e.} predicting the cognitive process from brain activity,
across many different studies touching different cognitive questions is a
key goal for cognitive neuroimaging as it provides a principled answer
to reverse inference~\cite{poldrack_decoding_2009-1}.
However, a major roadblock to scaling this approach is the necessity to
label cognitive tasks across studies in a rich but consistent way,
\emph{e.g.,} building an ontology~\cite{turner_cognitive_2012}.

We follow a more automated approach and cast dataset accumulation into a \textit{multi-task learning
problem}: our model is trained to decode simultaneously different
datasets, using
a shared architecture.
Machine-learning techniques can indeed
learn universal representations of inputs that give good
performance in multiple supervised problems~\cite{ando_framework_2005, xue_multi-task_2007}.
They have been successful, especially with the development of deep
neural network~\citep[see, \textit{e.g.,}][]{lecun_deep_2015}, in \textit{sharing
representations} and \textit{transferring knowledge} from one dataset prediction model to
another (\textit{e.g.,} in computer-vision \cite{donahue_decaf:_2014} and audio-processing \cite{collobert_unified_2008}). A popular approach
is to simultaneously learn to represent  the inputs of the different datasets in a low-dimensional space
and to predict the outputs from the low-dimensional representatives.
Using very deep model architectures in functional MRI is currently thwarted by the
signal-to-noise ratio of the available recordings and the relative little size of datasets
\cite{bzdok_inference_2017}
compared to computer vision and text corpora.
Yet, we show that multi-dataset representation learning is a
fertile ground for identifying cognitive systems with predictive power for mental operations.

\paragraph{Contribution.}
We introduce a new model architecture
dedicated to multi-dataset classification,
that performs two successive linear dimension reductions of the
input statistical brain images
and predicts psychological conditions from a \textit{learned}
low-dimensional representation of these images, linked to cognitive processes.
%
%
In contrast
to previous ontology-based approaches,
imposing a structure across different cognitive experiments is not needed
in our
model:
the
representation of brain images is learned using
the raw set of experimental conditions for each dataset. To our knowledge, this work
is the first to propose
knowledge aggregation and transfer learning in between functional MRI studies
with such modest level of supervision.
We demonstrate the performance of our model on several openly accessible and
rich reference datasets in the brain-imaging domain. The different aspects of
its architecture bring a substantial increase in out-of-sample accuracy
compared to models that forgo learning a cognitive-aware low-dimensional representation of brain maps. Our
model remains simple enough to be interpretable: it can be collapsed into a
collection of classification maps, while the space of low-dimensional
representatives can be explored to uncover a set of meaningful latent
components.

\section{Model: multi-dataset classification of brain statistical images}
\label{sec:model}
\begin{figure}
\centering
\includegraphics[width=\textwidth]{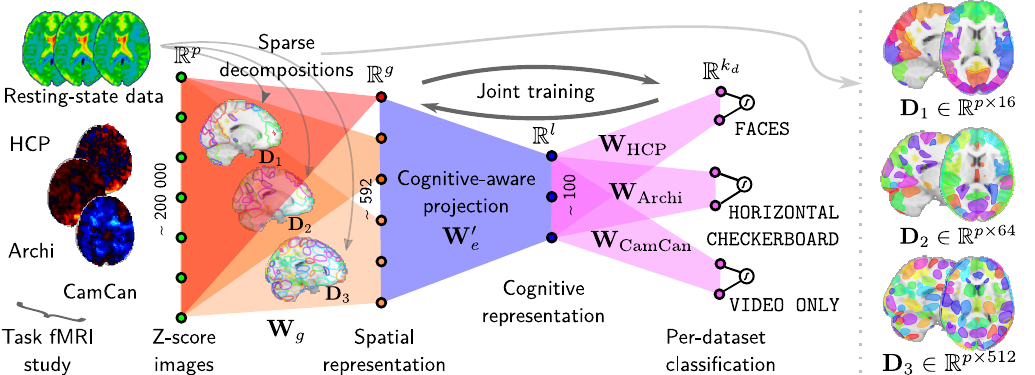}
\caption{\textbf{Model architecture: Three-layer multi-dataset classification.}
The first layer
(\textit{orange}) is learned from data acquired outside of cognitive experiments and captures a spatially coherent signal \textit{at multiple scales},
the second layer (\textit{blue}) embeds these representations
in a space common to all datasets, from which the conditions are predicted
(\textit{pink}) from multinomial models.}
\label{fig:abstract}
\end{figure}

Our general goal is to extract and
integrate biological knowledge across many brain-imaging studies
within the same statistical learning framework. We first outline how
analyzing large repositories of fMRI experiments can be cast as a
classification problem. Here, success in capturing
brain-behavior relationships is measured by out-of-sample
prediction accuracy. The proposed model (Figure~\ref{fig:abstract})
solves a range of these classification
problems in an identical statistical estimation and imposes a shared latent
structure across the single-dataset classification parameters.
These shared model parameters may be viewed as a chain of two dimension
reductions. The first reduction layer leverages knowledge about
brain spatial regularities; it is learned from resting-state data and designed
to capture neural activity patterns at different coarseness levels.
The second reduction layer projects
data on directions generally relevant for cognitive-state prediction. The
combination of both reductions yields low-dimensional
representatives that are less affected by noise and subject variance than the
high-dimensional samples: classification is expected to have better out-of-sample
prediction performance.


\subsection{Problem setting: predicting conditions from brain activity in
multiple studies}

We first introduce our notations and terminology, and
formalize a general prediction problem
applicable to any task fMRI dataset.
In a single fMRI \textit{study}, each subject performs different
\textit{experiments} in the scanner. During such an experiment, the subjects are
presented a set of sensory stimuli (\textit{i.e.}, \textit{conditions}) that aim at
recruiting a target set of cognitive processes.
We fit a first-level
general linear model for every record to obtain z-score maps that quantify the
importance of each condition in explaining each voxel. Formally, the $n$
statistical maps $(\x_i)_{i \in [n]}$ of a given study form a sequence in~$\RR^p$, where $p$ is the number of voxels in the brain. Each such observation
is labelled by a condition~$c_i$ in $[1,k]$ whose effect captures $\x_i$.
A single study typically features one or a few (if experiments are repeated) statistical map per condition
and per subject, and may present up to $k = 30$ conditions. Across
the studies, the observed brain maps can be modeled as generated from an
unknown joint distribution of brain activity and associated cognitive
conditions $((\x_i, c_i))_{i \in [n]}$ where variability across trials and
subjects acts as confounding noise. In this context, we wish to learn a
decoding model that predicts condition $c$ from brain activity $\x$ measured
from new subjects or new studies.

Inspired by recent work \cite{schwartz_mapping_2013,
bzdok_semi-supervised_2015, rubin_generalized_2016}, we frame the condition
prediction problem into the estimation of a multinomial classification model.
Our models estimate a
probability vector of $\x$ being labeled by each condition in $C$. This vector
is modeled as a function of~$(\W, \b)$ in $\RR^{p \times k} \times \RR^{k}$ that takes the \textit{softmax} form. For all $j$ in $[1,k]$, its $j$-th coordinate is defined as
\begin{equation}
    \p^{(j)}(\x, \W, \b) \triangleq \PP[c = j | \x, \W, \b]
       = \frac{e^{{\W^{(j)}}^\top \x + \b^{(j)}}}{\sum_{l \in C}
       e^{{\W^{(l)}}^\top \x + \b^{(l)}}}.
\end{equation}
Fitting the model weights is done
by minimizing the cross-entropy between ${(\p(\x_i))}_i$ and the true labels ${([c_i = j]_{j \in [k]})}_i$, with respect to
$(\W$,~$\b)$, with or without imposing parameter
regularization. In this model, an input image is classified in between all conditions presented in the whole \textit{study}. It is possible to restrict
 this classification to the set of conditions used in a given \textit{experiment} --- the empirical results of this work can be reproduced in this setting.

 \paragraph{The challenge of model parameter estimation.}A major inconvenience of the
 vanilla multinomial model lies in the ratio between the limited number of
 samples provided by a typical fMRI dataset and the overwhelming number of model weights
 to be estimated. Fitting the model amounts to estimating~$k$ discriminative
 brain map, \textit{i.e.} millions of parameters (4M for the 23 conditions of HCP), whereas most brain-imaging studies yield
 less than a hundred observations and therefore only a few thousands samples. This
 makes it hard to reasonably approximate the population parameters for
successful generalization, especially because the variance between subjects is
 high compared to the variance between conditions. The obstacle is
 usually tackled in one of two major ways in brain-imaging: 1) we can
 impose sparsity or a-priori structure over the model weights. Alternatively, 2) we can
 reduce the dimension of input data by performing spatial clustering or
 univariate feature selection by ANOVA.
 %
However, we note that, on the one hand, regularization strategies frequently incur
costly computational budgets if one wants to obtain interpretable
weights~\cite{gramfort_identifying_2013} and they introduce artificial bias. On
the other hand, existing techniques developed in fMRI analysis for dimension
reduction can lead to distorted signal and accuracy
losses~\cite{thirion_which_2014}.
Most importantly, previous statistical approaches are not tuned to
identifying conditions from task fMRI data. We therefore propose to use a dimension reduction
that is \textit{estimated} from data and tuned to capture the common hidden aspects
shared by statistical maps across studies --- we aggregate several classification models that share parameters.
%
\subsection{Learning shared representation across studies for decoding}\label{sec:shared}
We now consider several fMRI studies. ${(\x_i)}_{i \in [n]}$ is the union of all statistical maps from
all datasets. We write $D$ the set of all studies, $C_d$ the set of all $k_d$
conditions from study~$d$, $k \triangleq \sum_{d} k_d$ the total number of conditions and $S_d$ the subset of~$[n]$ that index samples
of study~$d$. For each study $d$, we estimate the parameters $(\W_d, \b_d)$ for the classification problem defined above.
Adapting the multi-task learning framework of~\cite{ando_framework_2005}, we constrain the
weights ${(\W_d)}_{d}$ to share a common latent structure: namely, we fix a latent dimension
$l \leq p$, and enforce that for all datasets $d$,
\begin{equation}\label{eq:factorization}
    \W_d = \W_e \W'_d,
\end{equation}
where the matrix $\W_e$ in $\RR^{p \times l}$ is shared across datasets,
and ${(\W'_d)}_d$ are dataset-specific classification matrices from a $l$ dimensional input space.
Intuitively, $\W_e$ should
be a ``consensus'' projection matrix, that project every sample
$\x_i$ from every dataset onto a lower dimensional representation $\W_e^\top
\x_i$ in $\RR^l$ that is easy to
label correctly.

The latent dimension $l$ may be chosen larger than $k$. In this case, regularization is necessary to ensure that the factorization~\eqref{eq:factorization} is indeed useful, \textit{i.e.}, that the multi-dataset classification problem does not reduce to separate multinomial regressions on each dataset. To regularize our model,
we apply \textit{Dropout}~\cite{srivastava_dropout:_2014} to the projected data representation.
Namely, during successive training iterations, we set a random fraction $r$ of the reduced data features to $0$. This prevents the co-adaptation of matrices $\W_e$ and ${(\W'_d)}_d$ and ensures that every direction of $\W_e$ is useful for classifying every dataset. Formally, Dropout amounts to sample binary diagonal matrices $\M$ in $\RR^{l \times l}$ during training, with Bernouilli distributed coefficients; for all datasets $d$, $\W'_d$ is estimated through the task of classifying Dropout-corrupted reduced data ${(\M \W_e^\top \x_i)}_{i \in S_d, \M \sim \mathcal{M}}$.

In practice, matrices $\W_e$ and ${(\W'_d)}_d$ are learned by jointly minimizing the following expected risk, where the objective is the sum of each of single-study cross-entropies, averaged over Dropout noise:
\begin{equation}
    \label{eq:erm}
    \min_{
    \substack{
    \W_e\\
    {(\W'_d)}_d
    }}
    \sum_{d \in D} \frac{1}{|S_d|}
    \sum_{i \in S_d}
    \sum_{j \in C_d}
    \EE_\M\big[ -\delta_{j = c_i}\log {\p_d^{(j)}[ \x_i, \W_e \M \W'_d, \b_d]}]\big]
\end{equation}
Imposing a common
structure to the classification matrices $(\W_d)_d$ is natural as the classes to
be distinguished do share some common neural organization
--- brain maps have a correlated
spatial structure, while the psychological conditions of the diffent datasets may trigger shared
cognitive primitives underlying human cognition
\cite{rubin_generalized_2016, bzdok_semi-supervised_2015}.
With our design, we aim at learning a matrix $\W_e$ that captures
these common aspects and thus benefits the generalization performance of
\textit{all} the classifiers. As $\W_e$ is \textit{estimated} from data, brain
maps from one study are enriched by the maps from all the other
studies, even if the conditions to be classified are not shared among
studies. In so doing, our modeling approach allows \textit{transfer learning}
among all the classification tasks.

Unfortunately, estimators provided by solving~\eqref{eq:erm} may have limited generalization performance as $n$ remain relatively small $(\sim 20,000)$ compared to the number of parameters. We address this issue by performing
an initial dimension reduction that captures the spatial structure of brain
maps.

\subsection{Initial dimension reduction using localized rest-fMRI
activity patterns}
\label{sec:geometric}
The projection expressed by $\W_e$ ignores the signal structure of
statistical brain maps. Acknowledging this structure in commonly acquired brain
measurements should allow to reduce the dimensionality of data with little
signal loss, and possibly the additional benefit of a denoising effect. Several recent
studies~\cite{blumensath_spatially_2013} in the brain-imaging domain suggest to use fMRI data acquired in
experiment-free studies for such dimension reduction. For this reason, we introduce a first
reduction of dimension that is not estimated from statistical maps, but from
resting-state data.
Formally, we enforce $\W_e = \W_g \W'_e$, where $g > l$ ($g
\sim 300$), $\W_g \in \RR^{p \times g}$ and $\W'_e \in \RR^{g \times k}$.
Intuitively, the multiplication by matrix~$\W_g$ should summarize the spatial
distribution of brain maps, while multiplying by $\W'_e$, that is estimated
solving~\eqref{eq:erm}, should find low-dimensional representations able to capture cognitive features. $\W'_e$ is now of reasonable size ($g \times l \sim
15000$): solving~\eqref{eq:erm} should estimate parameters with better
generalization performance. Defining an appropriate matrix $\W_g$ is the purpose of the next paragaphs.

\paragraph{Resting-state decomposition.}The initial dimension reduction
determines the relative contribution of statistical brain maps over what is
commonly interpreted by neuroscience investigators as \textit{functional
networks}. We discover such macroscopical brain networks by performing a sparse
matrix factorization over the massive resting-state dataset provided in the
HCP900 release \cite{vanessen_human_2012}: such a decomposition technique,
described \textit{e.g.,} in~\cite{mensch_dictionary_2016, mensch_stochastic_2017} efficiently provides
(\emph{i.e.}, in the order of few hours) a given number of sparse spatial maps that
decompose the resting state signal with good reconstruction performance. That
is, it finds a \textit{sparse} and \textit{positive} matrix $\D$ in $\RR^{p
\times g}$ and loadings $\A$ in $\RR^{g \times m}$ such that the $m$
resting-state brain images $\X_{rs}$ in $\RR^{p \times m}$ are well
approximated by $\D \A$. $\D$ is this a set of slightly overlapping networks ---
each voxel belongs to at most two networks. To maximally preserve Euclidian
distance when performing the reduction, we perform an \textit{orthogonal}
projection, which amounts to setting $\W_g \triangleq \D (\D^\T \D)^{-1}$.
Replacing in~\eqref{eq:erm}, we obtain the reduced expected risk minimization
problem, where the input dimension is now the number $g$ of dictionary
components:
\begin{align}
    \label{eq:erm_reduced}
    \min_{
    \substack{\W'_e \in \RR^{g \times l}\\
    {(\W'_d)}_d
    }}
    \sum_{d \in D} \frac{1}{|S_d|} \sum_{i \in S_d}
    \sum_{j \in C_d}
     \EE_\M\big[-\delta_{j = c_i}{\log \p_d^{(j)}[ \W_g^\top  \x_i, \W'_e \M \W'_d, \b_d]}\big].
\end{align}

\paragraph{Multiscale projection.}Selecting the ``best'' number of brain
networks $q$ is an ill-posed problem~\cite{eickhoff_connectivity-based_2015}:
the size of functional networks that will prove relevant for condition
classification is unknown to the investigator. To address this issue, we
propose to reduce  high-resolution data $(\x_i)_i$ in a multi-scale fashion: we
initially extract $3$ sparse spatial \textit{dictionaries} $(\D_j)_{j \in[3]}$
with $16$, $64$ and~$512$ components respectively. Then, we project statistical
maps onto each of the dictionaries, and concatenate the loadings, in a process
analogous to projecting on an overcomplete dictionary in computer vision
\cite[\textit{e.g.,}][]{mallat_matching_1993}. This amounts to define the matrix $\W_g$ as the concatenation
\begin{equation}
    \label{eq:w_g}
        \W_g \triangleq [\D_1 (\D_1^\top \D_1)^{-1}\:
        \D_2 (\D_2^\top \D_2)^{-1}\:
        \D_3 (\D_3^\top \D_3)^{-1}] \in \RR^{p \times {(16 + 64 + 512)}}.
\end{equation}
With this definition, the reduced data~${(\W_g^\top \x_i)}_i$ carry information
about the network activations at different scales. As such, it makes the
classification maps learned by the model more regular than when using a
single-scale dictionary, and indeed yields more interpretable classification
maps. However, it only brings only a small improvement in term of predictive
accuracy, compared to using a simple dictionary of size $k = 512$. We further discuss multi-scale decomposition in Appendix~\ref{app:multiscale}.

\subsection{Training with stochastic gradient descent}
As illustrated in Figure~\ref{fig:abstract}, our model may be interpreted as a
three-layer neural network with linear activations and several read-out heads,
each corresponding to a specific dataset. The model can be trained using
stochastic gradient descent, following a previously employed alternated
training scheme  \cite{collobert_unified_2008}: we cycle through
datasets $d \in D$ and select, at each iteration, a mini-batch of samples
$(\x_i)_{i \in B}$, where $B \subset S_d$ has the same size for all datasets. We
perform a gradient step --- the
weights $\W'_d$, $\b_d$ and $\W'_e$ are updated, while the others are left
unchanged. The optimizer thus sees the same number of samples for each dataset,
and the expected stochastic gradient is the gradient of \eqref{eq:erm_reduced},
so that the empirical risk decreases in expectation
and we find a critical point of~\eqref{eq:erm_reduced}
asymptotically. We use the Adam solver~\cite{kingma_adam:_2014} as a flavor
of stochastic gradient descent, as it allows faster convergence.
\paragraph{Computational cost.}Training the model on projected data
${(\W_g^\top \x_i)}_i$ takes 10 minutes on a conventional single CPU machine
with an Intel Xeon 3.21Ghz. The initial step of computing the dictionaries
$(\D_1, \D_2, \D_3)$ from all HCP900 resting-state (4TB of data) records takes
5 hours using~\cite{mensch_stochastic_2017}, while transforming data from all
the studies with $\W_g$ projection takes around 1 hour. Adding a new dataset
with 30 subjects to our model and performing the joint training takes no more
than $20$ minutes. This is much less than the cost of fitting a
first-level GLM on this
dataset ($\sim1\textrm{h}$ per subject).
\section{Experiments}
\label{sec:result}
We characterize the behavior and performance of our model on several large,
publicly available brain-imaging datasets. First, to validate the relevance of
all the elements of our model, we perform an ablation study. It proves that the
multi-scale spatial dimension reduction and the use of multi-dataset
classification improves substancially classification performance, and suggests
that the proposed model captures a new interesting latent structure of brain
images. We further illustrate the effect of \textit{transfer learning}, by
systematically varying the number of subjects in a single dataset: we show how
multi-dataset learning helps mitigating the decrease in accuracy due to smaller
train size --- a result of much use for analysing cognitive experiments on
small cohorts. Finally, we illustrate the interpretability of our model and
show how the latent ``cognitive-space'' can be explored to uncover some
template brain maps associated with related conditions in different datasets.
\subsection{Datasets and tools}
\paragraph{Datasets.}Our experimental study features 5 publicly-available task
fMRI study. We use all resting-state records from the HCP900
release~\cite{vanessen_human_2012} to compute the sparse dictionaries that are
used in the first dimension reduction materialized by~$\W_g$. We succinctly
describe the conditions of each dataset --- we refer the reader to the original
publications for further details.
\begin{itemize}
    \item \textbf{HCP}: gambling, working memory, motor, language, social and relational
    tasks. 800 subjects.
    \item \textbf{Archi }~\cite{pinel_fast_2007}: localizer protocol, motor, social and relational
    task. 79 subjects.
    \item \textbf{Brainomics }~\cite{papadopoulos_orfanos_brainomics/localizer_2017}: localizer
    protocol. 98 subjects.
    \item \textbf{Camcan }~\cite{shafto_cambridge_2014}: audio-video task, with frequency
    variation. 606 subjects.
    \item \textbf{LA5c consortium }~\cite{poldrack_phenome-wide_2016}: task-switching, balloon
    analog risk taking, stop-signal and spatial working memory capacity tasks --- high-level tasks.
     200 subjects.
\end{itemize}
The last four datasets are \textit{target datasets}, on which we measure
out-of-sample prediction performance. The larger HCP dataset serves as a
\textit{knowledge transfering dataset}, which should boost these performance
when considered in the multi-dataset model. We register the task time-series in
the reference MNI space before fitting a general linear model (GLM) and
computing the maps (standardized by z-scoring) associated with each
\textit{base} condition --- no manual design of contrast is involved. More details on the pipeline used for z-map extraction is provided in Appendix~\ref{app:data}.

\paragraph{Tools.}We use pytorch
\footnote{\url{http://pytorch.org/}} to define and train the proposed
models, nilearn~\cite{abraham_machine_2014} to handle brain
datasets, along with scikit-learn~\cite{pedregosa_scikit-learn:_2011} to design the experimental pipelines. Sparse brain decompositions were
computed from the whole HCP900 resting-state data. The code for reproducing experiments is available at
\url{http://github.com/arthurmensch/cogspaces}. Our model involves a few
non-critical hyperparameters:  we use batches of size $256$, set the latent
dimension $l = 100$ and use a Dropout rate $r=0.75$  in the latent cognitive space ---
this value perform slightly better than $r=0.5$. We use a multi-scale dictionary with $16$, $64$ and $512$
components, as it yields the best quantitative and qualitative results.\footnote{Note that using only the $512$-components dictionary yields comparable predictive accuracy. Quantitatively, the multi-scale approach is beneficial when using dictionary with less components (\textit{e.g.}, $16$, $64$, $128$) --- see Appendix~\ref{app:multiscale} for a quantitative validation of the multi-scale approach.}. Finally, test accuracy is measured
on half of the subjects of each dataset, that are removed from training sets
beforehand. Benchmarks are repeated 20 times with random split folds to
estimate the variance in performance.

\subsection{Dimension reduction and transfer improves test
accuracy}\label{sec:ablation}
For the four benchmark studies, the proposed model brings between +1.3\% to
+13.4\% extra test accuracy compared to a simple multinomial classification.
To further quantify which aspects of the model improve performance, we perform an ablation
study: we measure the prediction accuracy of six models, from the simplest to
the most complete model described in Section~\ref{sec:model}. The first three
experiments study the effect of initial dimension reduction and regularization\footnote{For
these models, $\ell_2$ and Dropout regularization
parameter are estimated by nested cross-validation.}. The
last three experiments measure the performance of the proposed factored model,
and the effect of multi-dataset classification.
\begin{figure}
    \centering
    \includegraphics{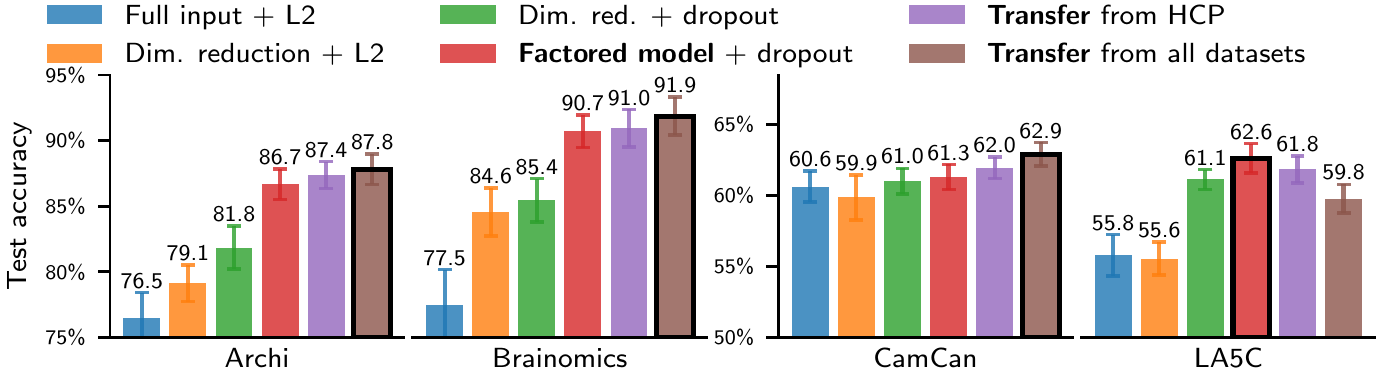}
    \caption{
    \textbf{Ablation results.} Each dimension reduction of the model has a
    relevant contribution. Dropout regularization is very effective when
    applied to the cognitive latent space. Learning this latent space allows to
    transfer knowledge between datasets.} \label{fig:ablation}
\end{figure}%
\begin{figure}
    \centering
    \includegraphics{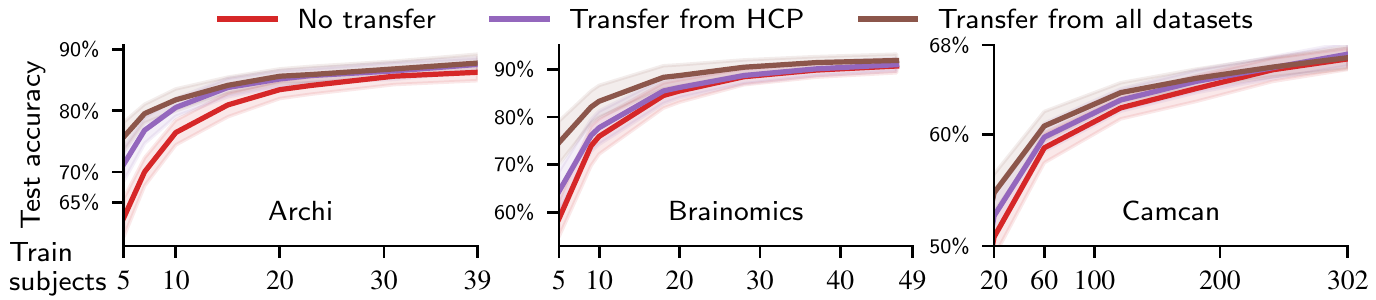}
    \caption{\textbf{Learning curves} in the single-dataset and multi-dataset setting.
    Estimating the latent cognitive space from multiple datasets is very useful
    for studying small cohorts.} \label{fig:learning_curve}
\end{figure}
\begin{enumerate}
    \item Baseline $\ell_2$-penalized multinomial classification, where we
    predict $c$ from $\x \in \RR^p$ directly.
    \item Multinomial classification after projection on a dictionary,
    \textit{i.e.} predicting $c$ from $\W_g \x$.
    \item Same as experience 2, using Dropout noise on projected data $\W_g \x$.
    \item Factored model in the single-study case:
    solving~\eqref{eq:erm_reduced} with the target study only.
    \item Factored model in a two-study case: using target study
    alongside HCP.
    \item Factored model in the \textit{multi}-study case: using target study
    alongside all other studies.
\end{enumerate}
The results are summarized in Figure~\ref{fig:ablation}. On average, both
dimension reduction introduced by~$\W_g$ and~$\W'_e$ are beneficial to
generalization performance. Using many datasets for prediction brings a further
increase in performance, providing evidence of transfer learning between
datasets.

In detail, the comparison between experiments 1, 2 and 3 confirms that
projecting brain images onto functional networks of interest is a good strategy
to capture cognitive information~\cite{bzdok_semi-supervised_2015,
blumensath_spatially_2013}. Note that in addition to improving the statistical
properties of the estimators, the projection reduces drastically the
computational complexity of training our full model. Experiment 2 and 3
measure the impact of the regularization method \textit{without} learning a
further latent projection. Using Dropout on the input space performs
consistently better than $\ell_2$ regularization ($\mathbf{+1\%}$ to $\mathbf{+5\%}$); this can
be explained in view of~\cite{wager_dropout_2013}, that interpret input-Dropout
as a $\ell_2$ regularization on the natural model parametrization.

Experiment 4 shows that Dropout regularization becomes much more powerful when learning a
second dimension reduction, \textit{i.e.} when solving
problem~\eqref{eq:erm_reduced}. Even when using a single study for learning, we
observe a significant improvement ($\mathbf{+3\%}$ to $\mathbf{+7\%}$) in performance on
three out of four datasets. Learning a latent space projection together with
Dropout-based data augmentation in this space is thus a much better
regularization strategy than a simple $\ell_2$ or input-Dropout regularization.

 Finally, the comparison between experiments 4, 5 and 6 exhibits the expected
 \textit{transfer} effect. On three out of four target studies, learning the
 projection matrix $\W'_e$ using several datasets leads to an accuracy gain
 from $\mathbf{+1.1\%}$ to $\mathbf{+1.6\%}$, consistent across folds. The more datasets
 are used, the higher the accuracy gain --- already note that this gain increases with smaller train size. Jointly classifying images on several
 datasets thus brings extra information to the cognitive model, which allows to
 find better representative brain maps for the target study. In particular, we
 conjecture that the large number of subjects in HCP helps modeling
 inter-subject noises. On the other hand, we observe a \textit{negative}
 transfer effect on LA5c, as the tasks of this dataset share little cognitive
 aspects with the tasks of the other datasets. This encourages us to use richer dataset repositories for further improvement.

\subsection{Transfer learning is very effective on small datasets}

To further demonstrate the benefits of the multi-dataset model, we vary the
size of target datasets (Archi, Brainomics and CamCan) and compare the
performance of the single-study model with the model that aggregates
Archi, Brainomics, CamCan and HCP studies.
Figure~\ref{fig:learning_curve} shows that the effect of transfer
learning increases as we reduce the training size of the target dataset.
This suggests that the learned data embedding $\W_g \W'_e$
does capture some universal cognitive information, and can be learned from
different data sources. As a consequence, aggregating a larger study to
mitigate the small number of training samples in the target dataset.
With only 5 subjects, the gain in accuracy due to
transfer is $\mathbf{+13\%}$ on Archi, $\mathbf{+8\%}$ on Brainomics, and $\mathbf{+6\%}$ on CamCan.
Multi-study learning should thus proves very useful to classify
conditions in studies with ten or so subjects, which are still very common in
neuroimaging.
\subsection{Introspecting classification maps}
\begin{figure}
    \centering
    \includegraphics{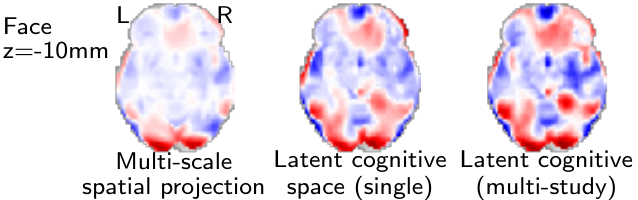}%
    \includegraphics{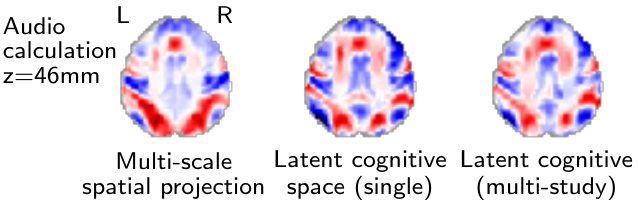}
    \caption{Classification maps from our model are more specific of higher level functions:
    they focus more on the FFA for faces, and on the \emph{left}
    intraparietal suci for calculations.}
    \label{fig:classification}
    \end{figure}
At prediction time, our multi-dataset model can be collapsed into one
multinomial model per dataset. Each dataset $d$ is then classified using matrix
$\W_g \W'_e \W'_d$. Similar to the linear models classically used for
decoding, the model weights for each condition can be represented as a brain
map. Figure~\ref{fig:classification} shows the maps associated with digit
computation and face viewing, for the Archi dataset. The models~2, 4 and 5 from the
ablation study are compared. Although it is hard to assess the intrinsic
quality of the maps, we can see that the introduction of the second projection
layer and the multi-study problem formulation (here, appending the HCP dataset)
yields maps with more weight on the high-level functional regions known to be
specific of the task: for face viewing, the FFA stands out more compared to
primary visual cortices; for calculations, the weights of the intraparietal
sulci becomes left lateralized, as it has been reported for symbolic number
processing~\cite{bugden_role_2012}.

\subsection{Exploring the latent space}
\begin{figure}
    \centering
    \includegraphics[width=.44\textwidth]{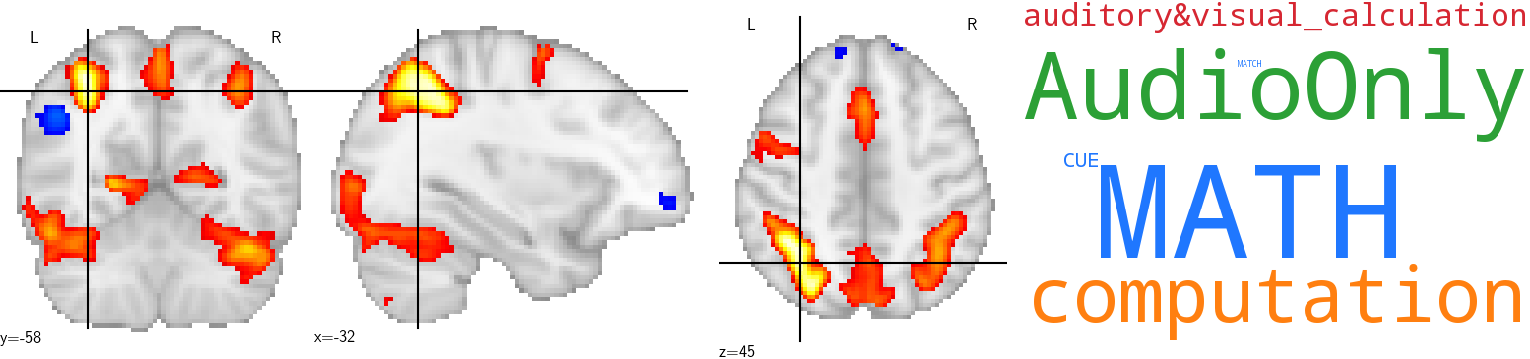}\hspace{.08\textwidth}%
    \includegraphics[width=.44\textwidth]{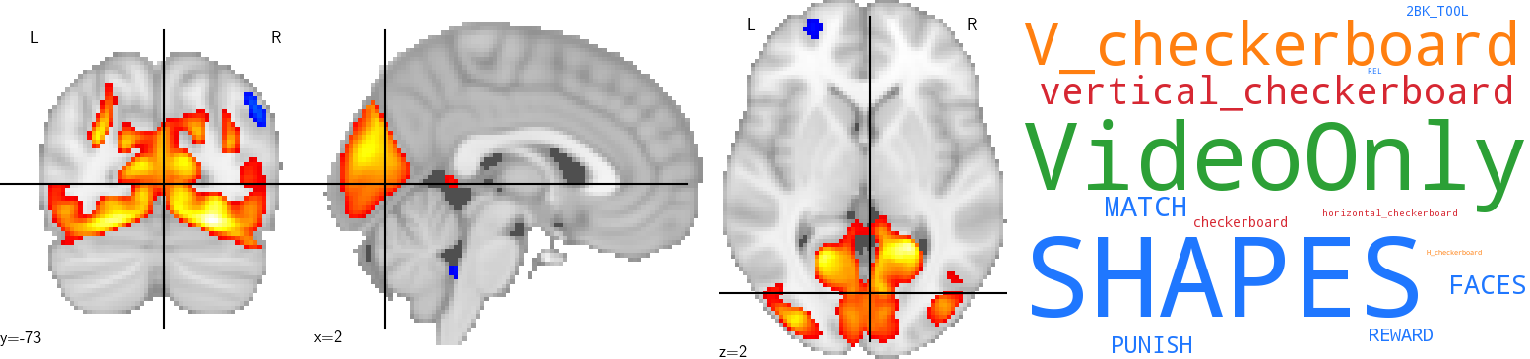}
    \includegraphics[width=.44\textwidth]{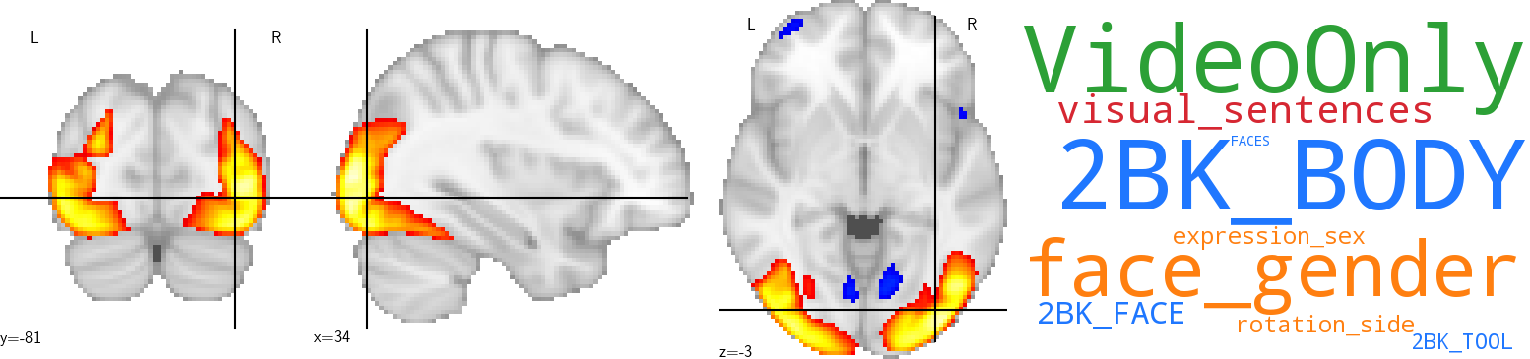}\hspace{.08\textwidth}%
    \includegraphics[width=.44\textwidth]{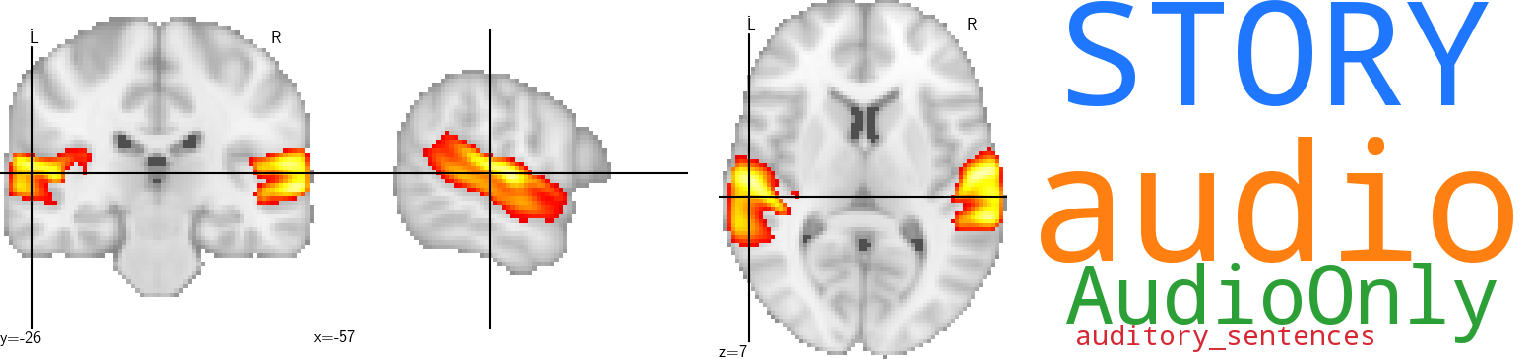}
    \includegraphics[width=.44\textwidth]{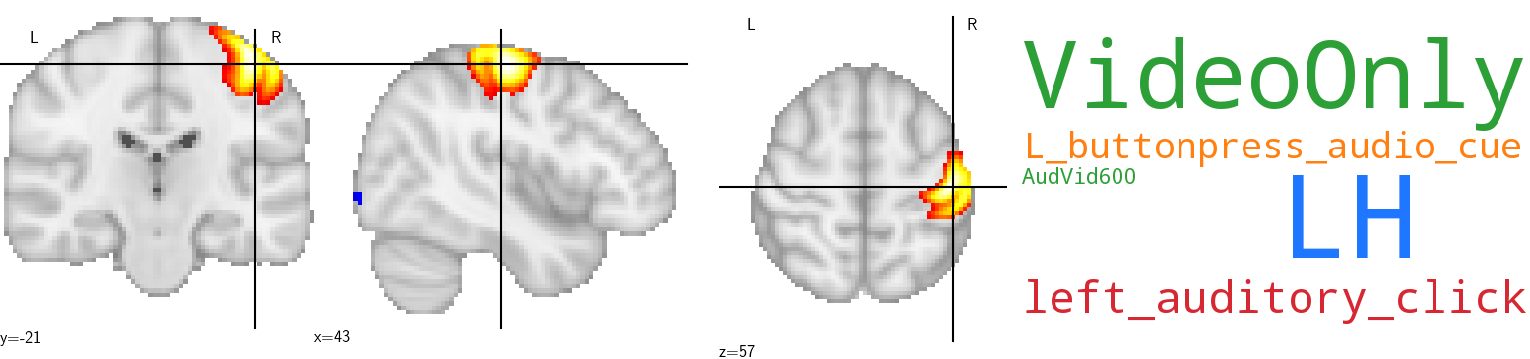}\hspace{.08\textwidth}%
    \includegraphics[width=.44\textwidth]{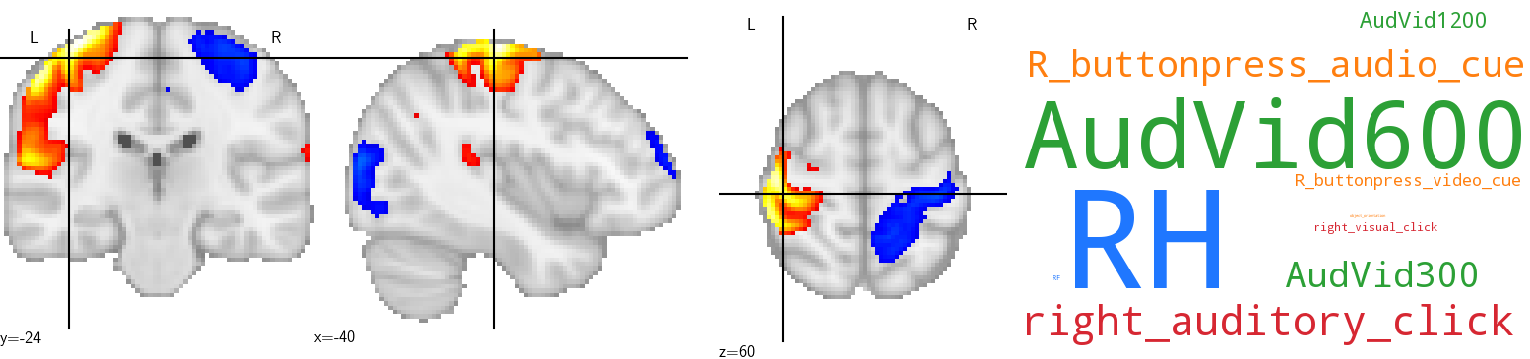}
    \caption{The latent space of our model can be explored to unveil some
    template brain statistical maps, that corresponds to bags of conditions
    related across color-coded datasets.}
    \label{fig:kmeans}
    \end{figure}

Within our model, classification is performed on the same $l$-dimensional space
$E$ for all datasets, that is learned during training. To further show that
this space captures some cognitive information, we extract from $E$
template brain images associated to general cognitive
concepts. Fitting our model on the
Archi, Brainomics, CamCan and HCP studies, we extract
representative vectors of $E$ with a k-means clustering over the projected
data and consider the centroids ${(\y_j)}_j$ of $50$ clusters.
Each centroid~$\y_j$ can be associated to a brain image
$\t_j \in \RR^p$ that lies in the span of $\D_1$,~$\D_2$ and~$\D_3$. In doing
so, we go backward through the model and obtain a representative of~$\y_j$ with well delineated spatial regions. Going forward, we compute the classification
probability vectors $\W_d^\top \y_j = {\W'_d}^\top {\W'}_e^\top \W_g^\top \t_j$ for
each study $d$. Together, these probability vectors give an indication on the cognitive
functions that $\t_j$ captures. Figure~\ref{fig:kmeans}
represents six template images, associated to their probability vectors,
shown as word clouds. We clearly obtain interpretable pairs of brain
image/cognitive concepts.
These pairs capture across datasets clusters of experiment conditions with
similar brain representations.

\section{Discussion}
We compare our model to a previously proposed formulation for brain image classification. We show how our model differs from convex multi-task learning,
and stress the importance of Dropout.
\paragraph{Task fMRI classification.}Our model is related to a previous semi-supervised classification
model~\cite{bzdok_semi-supervised_2015} that also performs multinomial
classification of conditions in a low-dimensional space: the dimension
reduction they propose is the equivalent of our projection $\W_g$. Our
approach differs in two aspects. First, we replace the initial
semi-supervised dimension reduction with unsupervised analysis of
resting-state, using a much more tractable approach that we have shown to be
conservative of cognitive signals. Second, we introduce the additional
cognitive-aware projection $\W'_e$, learned on multiple studies. It
substancially improves out-of-sample prediction performance, especially on
small datasets, and above all allow to uncover a cognitive-aware latent space, as we have
shown in our experiments.

\paragraph{Convex multi-task learning.}Due to the Dropout regularization and the fact that~$l$ is allowed to be
larger than $k$, our formulation differs from the classical
approach~\cite{srebro_maximum-margin_2004} to the multi-task problem,
that would estimate $\bTheta = \W'_e {[\W'_1, \dots, \W'_d]}_d \in \RR^{g \times k}$ by solving
a convex empirical risk minimization problem with a trace-norm penalization, that encourages $\bTheta$ to be low-rank.
We tested this formulation, which does not perform better than the explicit factorization formulation with Dropout regularization. Trace-norm regularized regression has the further drawback of being slower to train, as it typically operates with full gradients, \textit{e.g.} using FISTA~\cite{beck_fast_2009}. In contrast, the non-convex explicit factorization model is easily amenable to large-scale stochastic optimization --- hence our focus.

\paragraph{Importance of Dropout.}The use of Dropout regularization is crucial in our model. \textit{Without} Dropout,
in the single-study case with $l > k$, solving the factored problem~\eqref{eq:erm_reduced}
yields a solution worse in term of empirical risk than solving the simple
multinomial problem on ${(\W_g^\top \x_i)}_i$, which finds a global
minimizer of~\eqref{eq:erm_reduced}. Yet, Figure~\ref{fig:ablation} shows that
the model enriched with a latent space (\textit{red}) has better performance in test accuracy than the simple model (\textit{orange}), thanks to the Dropout noise applied
to the latent-space representation of the input data. Dropout is thus a
promising novel way of regularizing fMRI models.

\section{Conclusion}We proposed and characterized a novel cognitive
neuroimaging modeling scheme that blends latent factor discovery and transfer
learning. It can be applied to many different cognitive studies jointly without
requiring explicit correspondences between the cognitive tasks. The model helps
identifying the fundamental building blocks underlying the diversity of
cognitive processes that the human mind can realize. It produces a basis of
cognitive processes whose generalization power is validated quantitatively, and
extracts representations of brain activity that grounds the transfer of
knowledge from existing fMRI repositories to newly acquired task data. The
captured cognitive representations will improve as we provide the model with a
growing number of studies and cognitive conditions.

\section{Acknowledgments}

This project has received funding from the European Union's Horizon 2020 Framework Programme for Research and Innovation under grant agreement N\textsuperscript{o} 720270 (Human Brain Project SGA1). Julien Mairal was supported by the ERC grant SOLARIS (N\textsuperscript{o} 714381) and a grant from ANR (MACARON project ANR-14-CE23-0003-01). We thank Olivier Grisel for his most helpful insights.

\small
\putbib
\end{bibunit}

\appendix

\vfill
\pagebreak

\begin{bibunit}
\section{Supplementary material}

\subsection{Data processing pipeline}\label{app:data}

\paragraph{Resting-state fMRI.}To extract sparse dictionaries, we perform dictionary learning on resting
 state time-series from the 900 subject of the HCP900 dataset. These time
 series are already provided preprocessed and aligned in the MNI space. We apply a smoothing kernel of size 4\,mm to the data. We then use the
 modl\footnote{\url{http://github.com/arthurmensch/modl}} library to extract
 dictionaries from data by streaming records (in Nifti format) to an online
 algorithm. We run the algorithm on a grid of regularization parameters, that
 control the sparsity. We then select the sparsest dictionaries that cover all
 the brain. Dictionaries have been
 uploaded on our reproduction repository for potential
 further use.

\paragraph{Task fMRI.}To construct task fMRI datasets (HCP, Archi, Brainomics,
Camcan, LA5C), we perform as follow. We use SPM (via the pypreprocess
library\footnote{\url{http://github.com/neurospin/pypreprocess}}) for standard
preprocessing of task fMRI time series: motion correction, slice-time
interpolation and common registration to MNI space, using the DARTEL
algorithm~\cite{ashburner2007fast}. We use a smoothing kernel of size 4\,mm. We
then fit a first level general linear model to each subject. We use the result
of the regression to obtain one z-statistic map per base condition (a.k.a.
explatory variable) and per record. This pipeline can be streamlined for fMRI
datasets provided in BIDS~\cite{gorgolewski_brain_2016} format, using the
nistats library\footnote{\url{http://github.com/nistats/nistats}}.

\begin{figure}[h]
    \centering
    \includegraphics{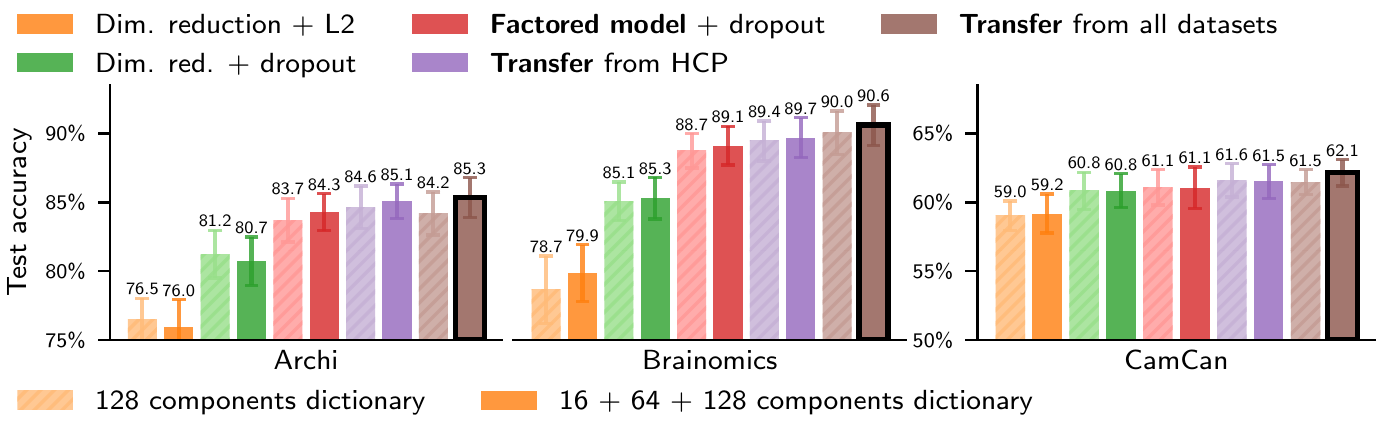}
    \caption{Using a multi-scale dictionaries consistently provide better performance than single-scale dictionaries for classical
     dictionary sizes.}
    \label{fig:multi-scale}
\end{figure}

\subsection{Multi-scale dictionaries}\label{app:multiscale}

To further investigate the interest of multi-scale projection on resting-state
dictionaries, we enforce a more agressive geometric dimension reduction in our
model than in the main text. That is, we use dictionaries with fewer components
than in Section~\ref{sec:result}. This is typically a way to improve
interpretability of the model by sacrifying a little out-of-sample accuracy. It
is also adviseable to restrain the size of the dictionary if ones use a smaller
resting-state dataset than HCP --- large dictionaries tend to overfit small
datasets.

In practice, we compare the dimension reduction using a $128$
components dictionary to the dimension reduction using the same dictionary,
along with a $16$ and a $64$ component dictionary. The results are presented in
Figure~\ref{fig:multi-scale}. We observe that multi-scale projection performs
systematically better than single scale projection. Note that the difference
between the performance of the two sets of model is significant as it
consistent across all folds. The multi-scale approach is thus quantitatively
useful when using relatively small dictionaries from resting state data.
\pagebreak
\renewcommand\refname{Extra references}
\putbib
\end{bibunit}
\end{document}